\documentclass[letterpaper, 10 pt, conference]{ieeeconf}  

\IEEEoverridecommandlockouts                              

\usepackage[l2tabu,orthodox]{nag}

\usepackage[
backend=biber, 
bibencoding=utf8,
style=ieee, 
sorting=none, 
natbib=true, 
doi=false, 
isbn=false, 
url=false, 
eprint=false, 
maxcitenames=1, 
mincitenames=1,
maxbibnames=5,
]{biblatex}

\usepackage[pdftex,colorlinks]{hyperref}

\usepackage[printonlyused]{acronym}
\acrodef{GNSS}{Global Navigation Satellite System}
\acrodef{RTK}{Real Time Kinematic}
\acrodef{ECU}{Electronic Control Unit}
\acrodef{MPC}{Model Predictive Control}
\acrodef{PID}{Proportional-Integral-Derivative}
\acrodef{IMU}{Inertial Motion Unit}
\acrodef{ANR}{French National Research Agency}
\acrodef{TIARA}{Toward Intelligent Adaptable Robots for Agriculture}
\acrodef{TSCF}{Technologies et systèmes d’information pour les agrosystèmes-Clermont-Ferrand}

\usepackage{siunitx}
\usepackage[all]{nowidow}

\usepackage[dvipsnames]{xcolor}

\usepackage{lipsum}

\usepackage[pdftex]{graphicx}
\usepackage{subcaption}
\usepackage{epstopdf}

\usepackage{import}

\graphicspath{{./latexGoodPractices/}}

\usepackage[dvipsnames]{xcolor}
\usepackage{tikz}
\usetikzlibrary{arrows, arrows.meta, calc, quotes,angles, patterns, intersections}

\usepackage{booktabs}

\usepackage{tabu}
\usepackage{soul}

\usepackage{amssymb,amsfonts,amsmath,amscd}

\usepackage{bm} 

\newcommand{\bbm}{\begin{bmatrix}}
\newcommand{\ebm}{\end{bmatrix}}

\usepackage{xspace}

\usepackage{fancyhdr}
\fancypagestyle{withfooter}{
  
  \fancyhead[L]{}
  \fancyhead[R]{}
  \fancyfoot[C]{\footnotesize Presented at the 2025 IEEE ICRA Workshop on Field Robotics}
}

\usepackage{microtype}
\usepackage{physics}

\title{\LARGE \bf
From Theory to Practice: Identifying the Optimal Approach for Offset Point Tracking in the Context of Agricultural Robotics
}

\author{Stephane Ngnepiepaye Wembe$^{1,2}$, Vincent Rousseau$^{1}$, Johann Laconte$^{1}$ and Roland Lenain$^{1}$
\thanks{$^{1}$Université Clermont Auvergne, INRAE, UR TSCF, 63000, Clermont-Ferrand, France; stephane.ngnepiepaye-wembe@inrae.fr}%
\thanks{$^{2}$SABI AGRI, 63360, Saint-Beauzire, France; stephane.ngnepiepaye-wembe@sabi-agri.com}%
}

\begin{document}

\maketitle
\thispagestyle{withfooter}
\pagestyle{withfooter}

\begin{abstract}
Modern agriculture faces escalating challenges: increasing demand for food, labor shortages, and the urgent need to reduce environmental impact. Agricultural robotics has emerged as a promising response to these pressures, enabling the automation of precise and sustainable field operations. In particular, robots equipped with implements for tasks such as weeding or sowing must interact delicately and accurately with the crops and soil. Unlike robots in other domains, these agricultural platforms typically use rigidly mounted implements, where the implement’s position is more critical than the robot’s center in determining task success. Yet, most control strategies in the literature focus on the vehicle body, often neglecting the actual working point of the system. This is particularly important when considering new agriculture practices where crops row are not necessarily straights. 
This paper presents a predictive control strategy targeting the implement’s reference point. The method improves tracking performance by anticipating the motion of the implement, which, due to its offset from the vehicle’s center of rotation, is prone to overshooting during turns if not properly accounted for.
\end{abstract}

\section{Introduction}
Agriculture faces the dual challenge of feeding a growing population while reducing pollution. 
Key concerns include preserving soil and ensuring sufficient food production. 
According to The State of Food Security and Nutrition in the World 2024, 733 million people experienced hunger in 2023, roughly one in eleven globally~\cite{fao_state_2024}. 
To come through this challenge, two dominant farming models coexist: conservation agriculture, which is environmentally sustainable but lower-yielding, and intensive agriculture, which increases output using pesticides and fertilizers but harms ecosystems. 
Historically, balancing productivity and environmental protection has involved trade-offs.

\begin{figure}[t] 
\centering 
\includegraphics[width=\linewidth]{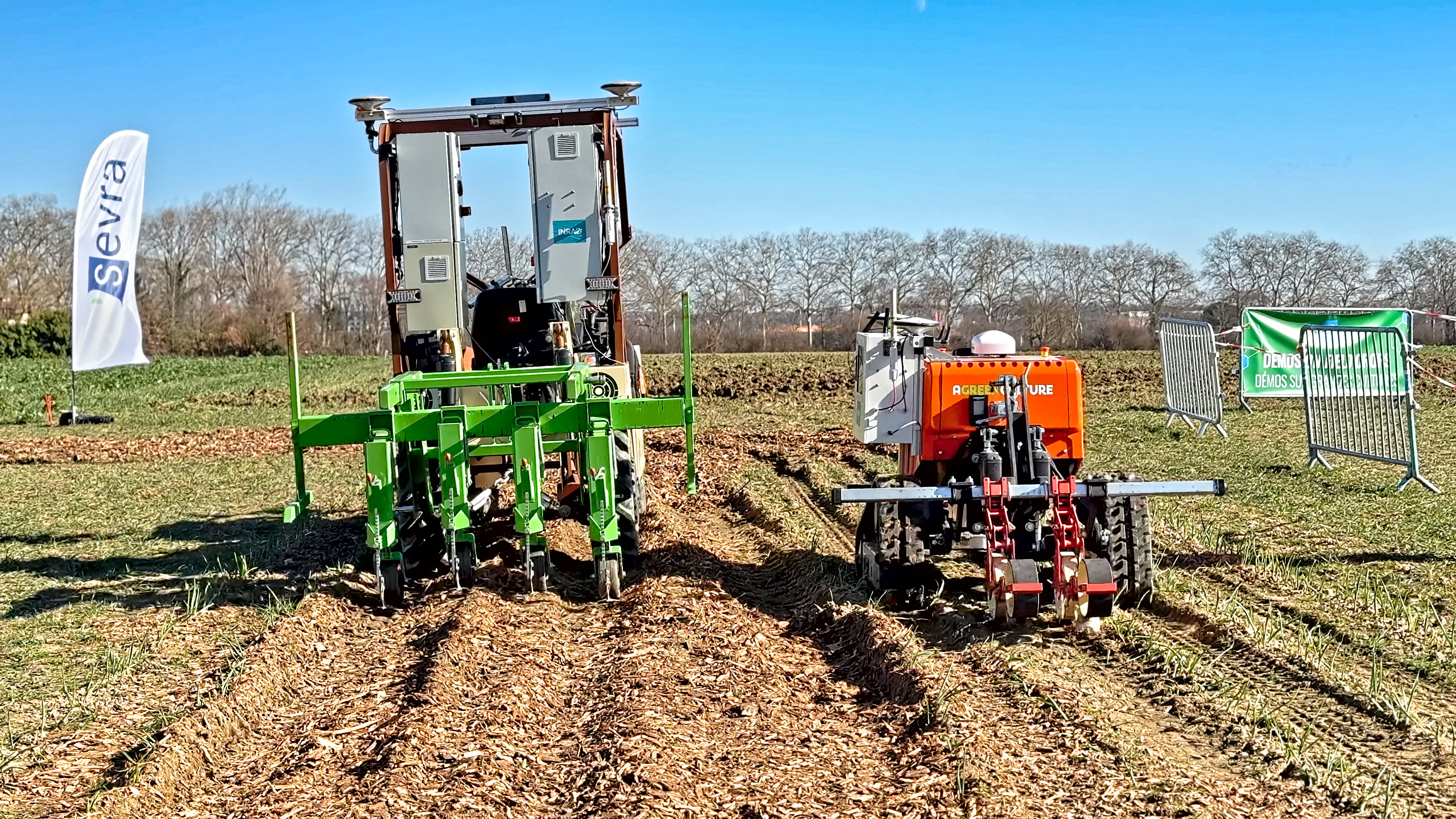} \caption{Agricultural robots with mounted implements. Precision is required at the implement-soil interface, making it essential to focus control on the implement rather than the robot’s center.} 
\label{fig:intro} 
\end{figure}

Recently, agroecology has gained traction, merging ecological principles with productivity goals. 
In this context, Agriculture 4.0 and robotics play a vital role~\cite{lenain2021agricultural}. 
Robotic systems automate tasks like planting and weeding, reducing chemical inputs and environmental damage~\cite{lenain_robotique_2019}. 
As robotics aligns closely with agroecology, research efforts increasingly focus on how robots can be effectively integrated into farms~\cite{inbook_Robotics_in_Agriculture}. 
Core robotic functions such as perception, localization, planning, and control are essential, especially for path-following, which is critical in precise crop management.
Accurate path following is vital to perform agricultural operations. 
A weeding implement, for example, must remove weeds without harming the plants, as illustrated in~\autoref{fig:intro}. 
Poor tracking performance risks crop damage, emphasizing the importance of refined control algorithms.

Although literature offers a range of path-following techniques from simple approaches like Pure Pursuit~\cite{Coulter-1992-13338} to advanced predictive models such as \ac{MPC}~\cite{article_PMC}, most strategies focus on the robot’s rear axle for simplicity. 
However, for agricultural tasks requiring high precision, the control focus should shift to the implement which is directly interact with crops, and which is often offset from the robot’s center.
To focus control on the implement, most studies address trailer systems with pivot joints, typically mounted at the rear~\cite{Cariou_trailer}. 
Yet, many agricultural implements use rigid connections, posing greater challenges for precision, especially in the case when it is attachecd at the rear of the tractor~\cite{gan2007implement}.

This paper is built on our earlier work~\cite{previous_work} and proposes a novel optimal control strategy for managing an offset point. The method is predictive and tailored to minimize tracking errors during curvature transitions. Our contributions are:

\begin{itemize} 
\item A new optimal predictive control law for offset-point path following;
\item Experimental validation through real-world agricultural field trials; and
\item An analysis of how the prediction horizon affects tracking performance. \end{itemize}

\section{Related Work}In this section, we present path-following approaches applied in the agricultural domain, emphasizing robot precision and focusing on control strategies described in the literature.

\subsection{Path Following Focused on the Robot’s Center}

Various methods have been developed to guide a robot’s center along a predefined path, including point-to-point navigation~\cite{article_point_to_point}, trajectory tracking with temporal constraints~\cite{inproceedings_Time_constrained_trajectory_tracking}, and path following without such constraints~\cite{path_following_based_on_position}. 
In agricultural contexts, path following is often preferred due to its adaptability to terrain and task-specific requirements. 
These algorithms are typically classified into four groups: geometric-based, kinematic-based, dynamic-based, and neural network-based methods.

Geometric control methods derive from spatial relationships between the robot and the path. 
The two most widely used algorithms in this category are Pure Pursuit and the Stanley controller. 
Originally developed for missile guidance~\cite{Scharf_pure_pursuit}, Pure Pursuit was later adapted for mobile robotics~\cite{Coulter-1992-13338}. 
It determines a target point ahead on the path using a “look-ahead-distance,” then steers the robot in an arc toward it. 
While simple and popular, improper tuning can cause oscillations because it does not explicitly consider the robot's orientation.
The Stanley controller, developed for the DARPA Grand Challenge~\cite{thrun_stanley_2007}, directly incorporates vehicle orientation into its control law, eliminating the need for the look-ahead-distance parameter. 
Despite this, it is sensitive to actuator delays, particularly in dynamic environments~\cite{Hoffmann_standley}. 
Overall, while geometric methods are easy to implement, their reliance on spatial heuristics may limit movement smoothness.

Kinematic-based methods use the robot’s motion characteristics such as velocities and accelerations to improve tracking precision~\cite{rokonuzzaman_review_2021}. 
These include feedback, predictive, and adaptive controls. 
Feedback control involves linearizing the system, allowing for~\ac{PID} controllers. 
This can be approximate or exact, depending on the system~\cite{Samson_chained}. 
Though simplifying controller design, linearization may limit generality.
Adaptive strategies like backstepping~\cite{Kokotovic_backstepping} improve stability for nonlinear systems but require accurate models and are sensitive to disturbances. 
Among predictive techniques,~\ac{MPC} is prominent. It anticipates future states to optimize actions under constraints~\cite{ding_model_2018}. For instance,~\cite{Soitinaho_MPC} proposes an MPC controller that minimizes tracking error, effort, and obstacle interaction. 
However, ~\ac{MPC} is computationally intensive and often lack convergence guarantees. 

Dynamic-based methods address factors like slippage, common in agricultural environments. 
By modeling such effects~\cite{Tazzari_slip}, more accurate controllers can be developed. 
For example,~\cite{Eaton_sliding_backstepping} integrates steering dynamics into a backstepping controller, improving performance under disturbance. 
Yet, these models increase system complexity and nonlinearity.

Finally, neural network-based methods offer flexible solutions for nonlinear, uncertain conditions. They dynamically adjust control gains and handle effects like slippage~\cite{article_NN_PID}. One example is a neural PID controller that tunes itself in real time~\cite{article_NN_PID}. While effective, these methods demand extensive training and validation to ensure reliability.

\subsection{Control Strategies for Agricultural Implements}

Given that agricultural robots mainly serve as implement carriers, control strategies most prioritize the trajectory of the implement rather than the robot itself. The objective is to ensure precise tracking of the implement’s path. 
Studies on control laws focused on the implement have recently emerged. These approaches can be grouped according to the type of connection between the robot and the implement.

Trailer systems, where the implements are attached via pivot joints which can be passive or active. 
Control approaches are used to enable the trailer system to follow a predefined path:
\citet{backman_nonlinear_2010} propose an active control strategy based on Nonlinear~\ac{MPC} that achieves high-precision. 
The presence of pivot joints proves especially beneficial during turning maneuvers, where they help minimize lateral errors.

In contrast, when implements are rigidly connected, particularly at the rear, significant lateral tracking errors have been reported~\cite{gan2007implement}. To address this, the tractor and implement were modeled as decoupled systems by \citet{Freimann2007ABA}, although their mutual influence was not considered.
A more integrated approach was proposed in~\cite{lukassek_model_2020}, where a~\ac{MPC} strategy was used to constrain a point on the implement to a reference trajectory. However, this method was limited to front-mounted implements and lacked experimental validation.
Previously, two strategies were proposed to regulate an offset point, either laterally or longitudinally positioned~\cite{previous_work}. The first relied on lateral error servoing, while the second adopted a backstepping design. Both were found to be sensitive to disturbances such as curvature changes, thus reducing overall robustness.
In this study, an optimal predictive control framework is proposed to regulate a point rigidly attached to either end of the tractor. Experimental validations are also conducted to evaluate its effectiveness.

\section{Preliminaries}
In this section, we present a novel approach for controlling an offset point on the robot, such as an agricultural implement.
We describe the assumptions and robot modeling and we introduce an optimal control approach designed to overcome the drawbacks identified in our previous method~\cite{previous_work}.

\subsection{Modeling} 
For the purpose of modeling, the following assumptions are made:

\begin{description} 
    \item[H1] The robot operates on a surface where all wheels remain in continuous contact with the ground. 
    \item[H2] Dynamic effects are assumed to be negligible.
    \item[H3] Wheel slip is considered negligible. 
    \item[H4] The offset point (e.g., the implement) is rigidly connected to the robot. 
    \item[H5] The implement’s lateral offset remains smaller than the minimum radius of curvature of the path.
    \item[H6] The robot is symmetric with respect to the vertical sagittal plane passing through the center of the rear axle.
\end{description}

Assumptions H1 to H6 are identical to those made in our previous work~\cite{previous_work}, where they are thoroughly justified.

\begin{figure}[htbt]
    \centering
    \includegraphics[width=\linewidth]{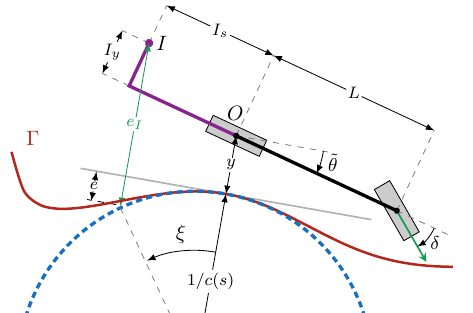}
    \caption{Notation used in this paper. The robot is modeled using a bicycle model, with $O$ as its center. Unlike conventional approaches aligning $O$ to the path, control is applied to an offset point $I$, representing the implement, to ensure accurate path following.}
    \label{fig:model}
\end{figure}

\autoref{fig:model} illustrates the notation used in modeling the robot. 
As previously mentioned, hypothesis H6 enables modeling the robot using a bicycle model. 
The main difference in this approach is that the control law does not focus on the robot's center, but rather on the implement's offset point, denoted as $I$. 
Unlike traditional approaches, which consider the robot's center, here the control explicitly considers the implement's reference point $I$.

The implement's control point is defined by offsets $I_s$ and $I_y$ relative to the robot frame centered at $O$, which can be positive or negative, allowing flexible placement around the robot. 
The angular deviation $\tilde{\theta}$ and lateral error $y$ at the rear axle center are measured via \ac{GNSS} relative to the geo-referenced path $\Gamma$, locally approximated as circular with curvature $c(s)$. 
The objective is to regulate the implement’s lateral deviation $e_I$, defined as its closest distance to $\Gamma$, measured along the $y$-axis direction.

\subsection{Kinematic Model of the offset Point I}
In order to directly regulate the position of the implement $I$, it is necessary to have a kinematic model of the state variables. As presented in our previous work~\cite{previous_work}, the kinematic of the implement lateral deviation with respect to time is written as
\begin{equation}
    \begin{aligned}
		\dot{e}_I & = v\sin\tilde{\theta} + \dot{\tilde{\theta}} \left(I_s  \cos\tilde{\theta} - I_y \sin\tilde{\theta}\right),
	\end{aligned} 
 \label{eq:DotYT}
\end{equation}
where $\dot{e}_I$ denoted the derivative with respect to time.
Finally, following the formulation in~\cite{Lenain2004AdaptiveAP} and assuming negligible slip, the vehicle’s kinematic model is expressed as:
\begin{equation}
    \begin{aligned}
        \dot{s} &= \frac{v\cos{\tilde{\theta}}}{1-c(s)y}, \text{and}\\
        \dot{\tilde{\theta}}& = v\left(\frac{\tan(\delta)}{L} - \frac{c(s)\cos\tilde{\theta}}{1-c(s)y} \right), 
    \end{aligned} 
    \label{eq:DotTheata}
\end{equation}
where $v$ is the vehicle velocity.
Both quantities remain well-defined as long as $y \neq 1/c(s)$, since positioning the robot at the center of the osculating circle results in a singularity. Nevertheless, such configurations are unlikely to occur in practice, as the curvature typically remains low due to steering limitations.


\section{Control of the Offset Point $I$}
In our previous work~\cite{previous_work}, we presented two approaches for controlling an offset point. Although both methods achieved steady-state convergence, we observed significant deviations during transitional phases. In this section, Using the kinematics of the control point defined above, we introduce an optimal approach aimed at minimizing these deviations during transitions state.

A backstepping-based control law, combined with a~\ac{MPC} strategy inspired by the approach developed in~\cite{picard_predictive}, is proposed to directly regulate the implement’s lateral deviation $e_I$ via the steering angle $\delta$. The proposed method is organized in two stages:
(1) an optimal desired orientation $\tilde{\theta}_d^h$ is computed through a~\ac{MPC} formulation to ensure exponential convergence of $e_I$; and
(2) a control law is derived to adjust the steering angle $\delta$ so as to track the reference orientation $\tilde{\theta}_d^h$.

For practical applications in the agricultural domain, analyzing the convergence properties of the control laws in terms of distance rather than time proves to be more suitable.

In the first stage, the optimal desired orientation $\tilde{\theta}_d^h$ is computed to ensure exponential convergence of the implement $I$ toward the reference path. As a result, the derivative $\dot{\tilde{\theta}}$ is not directly controlled, but is assumed to be measurable and is denoted as $\bar\omega$. Accordingly, the kinematic model of the implement’s lateral deviation $e_I$ is reformulated with respect to the curvilinear abscissa as follows:

\begin{equation} 
    \begin{aligned} 
        e'_I &= \alpha\left[\tan\tilde\theta+ \gamma\left(I_s - I_y \tan\tilde\theta\right)\right], 
    \end{aligned} 
    \label{eq:DerivYT} 
\end{equation} 
where 
\begin{equation} 
    \begin{aligned} 
        \alpha &= 1 - c(s)y, \quad\text{and } \gamma = \frac{\bar\omega}{v}. 
    \end{aligned} 
    \label{eq:defalphagamma} 
\end{equation}

As expected, when expressing the lateral error $e_I$ in terms of distance, it is necessary to assume that the robot's forward velocity $v$ remains strictly positive ($v>0$).

Neglecting the yaw acceleration $\ddot{\tilde\theta}$ and the term involving $\dot{\tilde\theta}^2$, which are assumed to be negligible due to the small yaw rate, the second-order dynamics of the implement’s lateral deviation can be written as:

\begin{equation} 
    \begin{aligned} 
        e''_I &= \dv{e_I}{t^2}\left(\dv{s}{t}\right)^{-2} \\
            &= \frac{\alpha^2}{\cos\tilde{\theta}} \left[\frac{\tan(\delta)}{L} - \frac{c \cos\tilde\theta}{\alpha}\right], 
    \end{aligned} 
    \label{eq:DerivDerivYT} 
\end{equation}

~\autoref{eq:DerivYT} and~\autoref{eq:DerivDerivYT} are used in the first stage of the proposed control method.

\subsection{First stage: A~\ac{MPC}-based determination of the vehicle's optimal desired orientation}

In this first stage, the optimal desired orientation of the robot is determined so that exponential convergence of the implement point $I$ toward the reference path is ensured.
Let $s_h$ denote the curvilinear abscissa associated with a prediction horizon  $h$. Under this assumption, and by neglecting terms of order $O(\Delta s^3)$, the lateral deviation at position $s + \Delta s$, with $\Delta s \in [0, s_h]$, is expressed as:

\begin{equation}
    \begin{aligned} 
        e_I(s+\Delta s) = e_I(s)+ e_I'(s)\Delta s + e_I''(s) \Delta s^2. 
    \end{aligned} 
    \label{eq:ytfutur} 
\end{equation}

In this formulation, the term $e_I'(s)\Delta s$ represents the deviation resulting from direction changes over the horizon, while $e_I''(s) \Delta s^2$ accounts for the effect of curvature variation.

The prediction horizon is discretized into $n_h \in \mathbb{N}$ points, which are linearly spaced over the interval $[0, s_h]$, leading to a sampling step of $s_t = s_h / n_h$.
Accordingly, the lateral deviation at the $k^{\text{th}}$ sampling point (where $k \in [0, n_h]$ and $\Delta s = ks_t$) is given by:
\begin{equation} 
    \begin{aligned} 
        e_I^k = e_I(s) + e_I'(s) (ks_t) + e''_I(ks_t)^2. 
    \end{aligned} 
    \label{eq:PredictiveControlLaw2} 
\end{equation}
In this expression, $e_I'(s)$ is defined according to~\autoref{eq:DerivYT}, and $e''_I$ is assumed to remain constant over the prediction horizon. Its value is computed using~\autoref{eq:DerivDerivYT} and is taken as the one corresponding to the end of the prediction horizon. This enables anticipation with respect to curvature changes.

The optimal predictive control is obtained by minimizing the sum of the lateral deviations $e_I^k$ over the horizon ($0 < k \leq n_h$). To this end, the following cost function is considered:
\begin{equation} 
    \begin{aligned} 
      J = \sum_{k=1}^{n_h}\left[e_I^k - e_I(s)e^{-\lambda ks_t} \right]^2 .
    \end{aligned} 
    \label{eq:optimisatiocriterium} 
\end{equation}
The term $e_I(s)e^{-\lambda ks_t}$ defines the desired exponential convergence profile for the lateral deviation $e_I$. This convergence behavior is shaped through the parameter $\lambda$. As such, the criterion $J$ quantifies the deviation between the predicted implement error and the reference convergence path.

By expanding~\autoref{eq:optimisatiocriterium}, the following expression is obtained:
\begin{equation} 
    \begin{aligned} 
        J = \sum_{k=1}^{n_h}\left[e_I + \xi ks_t + \alpha \gamma I_s(ks_t) + e''_I(ks_t)^2 - e_I e^{-\lambda ks_t}\right]^2. 
    \end{aligned} 
    \label{eq:optimisatiocriterium2} 
\end{equation}
With the following change of variable:

\begin{equation} 
    \begin{aligned} 
        \xi = \alpha (1-\gamma I_y)\tan\tilde\theta, 
    \end{aligned} 
    \label{eq:changementvariable} 
\end{equation}
the optimal condition is obtained by setting the partial derivative of $J$ with respect to $\xi$ to zero. The value of $\xi$ that minimizes the criterion is denoted $\xi_d^h$ and is given by:
\begin{equation} 
    \begin{aligned} 
        \xi_d^h = - \frac{1}{\sigma_2} \left[e_I(s)\sigma_1 + \alpha \gamma I_s\sigma_2 + e''_I \sigma_3 - e_I(s)\sigma_e\right], 
    \end{aligned} 
    \label{eq:PredictiveControlLaw3} 
\end{equation}

where the following quantities are defined:
\begin{equation}
    \begin{aligned}
        \left\{
            \begin{array}{llll}
                \sigma_1 = \sum_{k=0}^{n_h} (ks_t)  \\
                \sigma_2 = \sum_{k=0}^{n_h} (ks_t)^2  \\
                \sigma_3 = \sum_{k=0}^{n_h} (ks_t)^3 \\
                \sigma_e = \sum_{k=0}^{n_h} (ks_te^{-\lambda.ks_t}) .\\
                
                \end{array}
        \right.
    \end{aligned} \label{eq:PredictiveControlLaw3}
\end{equation}

The desired angular deviation $\tilde\theta_d^h$ is then recovered using:

\begin{equation} \begin{aligned} \tilde\theta_d^h = \arctan\left(\frac{\xi_d^h}{\alpha(1-\gamma I_y)}\right) \end{aligned} \label{eq:PredictiveControlLaw} \end{equation}

Two singularities are observed in this formulation: one occurs when $y = 1/c(s)$, as noted in~\autoref{eq:DotTheata}, and the other when $1 - \gamma I_y = 0$. However, as previously discussed in~\cite{previous_work}, such conditions are unlikely in practice, given that the implement generally remains close to the vehicle.

When the robot’s angular deviation matches the desired value $\tilde\theta_d^h$, the optimal condition is fulfilled, and the implement’s lateral deviation is driven to converge. Furthermore, since the cost function $J$ is convex, the resulting trajectory ensures a minimal cumulative lateral error during convergence.

Accordingly, the remaining task consists in ensuring rapid convergence of the robot’s angular deviation $\tilde\theta$ toward the desired value $\tilde\theta_d^h$.

\subsection{Second Stage: compute the steering angle}
For the second backstepping stage, we define the error $e_\theta=\tilde\theta-\tilde\theta_d^h$, and neglecting the variations of $\tilde\theta_d^h$, As demonstrated in our previow work~\cite{previous_work}, the steering angle that ensures the convergence of the angular deviation to $\tilde\theta_d^h$ is
\begin{equation}
\begin{aligned}
		\delta^d & = \arctan (L\frac{\left[-k_\theta e_\theta + c(s) \right]\cos{\tilde\theta}}{(1-c(s)y)}).
\end{aligned} \label{eq:ControlLawB}
\end{equation}
In conclusion, given the lateral and angular deviations $y,\tilde\theta$ measured by on-board sensors, the implement error $e_I$ can be computed~\cite{previous_work}. 
From this, the optimal current desired angular deviation $\tilde\theta_d^h$ is determined with \autoref{eq:PredictiveControlLaw}, and the steering angle to apply to the robot is finally found with \autoref{eq:ControlLawB}, driving the implement toward the path.

This approach combines the advantages of backstepping control, such as the regulation of intermediate variables like the vehicle's orientation, with a predictive aspect that allows it to anticipate changes in curvature along the reference path. Furthermore, its optimal nature ensures the minimization of implement lateral deviations even during transient state.

\section{Experiments}
In this section, an analysis of the proposed method is provided in a real-world scenario.
To this end, a comparison is conducted with existing offset point servoing methods found in the literature. Three approaches have been identified. The first is based on a~\ac{MPC} strategy, as presented in~\cite{lukassek_model_2020}. The second relies on lateral control of the robot’s center, while the third follows a reactive strategy based on backstepping, both presented in~\cite{previous_work}.
The first approach was designed for implements positioned exclusively at the front of the vehicle, and no experimental validation was provided, thereby limiting its relevance as a baseline for comparison. Moreover, since the implementation code was not provided, this approach is currently being implemented and will be compared with the proposed method in a forthcoming paper.
As a result, the proposed approach is compared with the two methods developed in~\cite{previous_work}.

Two main experiments were conducted. In the first, a comparison of the proposed approach is carried out on a relatively simple path, consisting of a straight segment, a positively curved segment, and a negatively curved segment, achieved on a grass ground (see~\autoref{fig:traj01}).
With this experiment, the validity of the optimal approach is demonstrated by showing that, similarly to the methods developed in~\cite{previous_work}, it enables the implement $I$ to follow a reference path.
Throughout the transient phases of the experiment, the robustness of the optimal approach is highlighted, as it significantly reduces implement deviations during curvature transitions.

Furthermore, since the proposed method is inherently predictive, it benefits from a prediction horizon that helps mitigate the effects of disturbances. To evaluate the influence of this prediction horizon, the second experiment investigates its impact on the lateral error of the implement when following a more complex reference path subject to multiple disturbances (as depicted on the ~\autoref{fig:trajries}). This experiment enables a more detailed analysis of how the prediction horizon affects tracking accuracy along the path.

\subsection{Experimental setup}
The following experiments were carried out using the robot depicted on the left in~\autoref{fig:intro}.
The platform is a four-wheel-drive, Ackermann-type vehicle capable of towing various implements, such as mechanical weeders.
It is equipped with an \ac{RTK} \ac{GNSS} sensor, which is used to compute the lateral and angular deviations $y$ and $\tilde\theta$ relative to a geo-referenced, pre-recorded path.
The term $\gamma = \frac{\bar\omega}{v}$, used in the first stage of the method, is obtained by measuring the steering angle $\delta$ with an encoder sensor. The value $\bar\omega$ is then computed using~\autoref{eq:DotTheata} and the measured steering angle.
All experiments were conducted at a speed of $v = \SI{1.0}{\m/\s}$.

\subsection{Validation of  the method }
\begin{figure}[thbp]
\centering
\includegraphics[width=\linewidth]{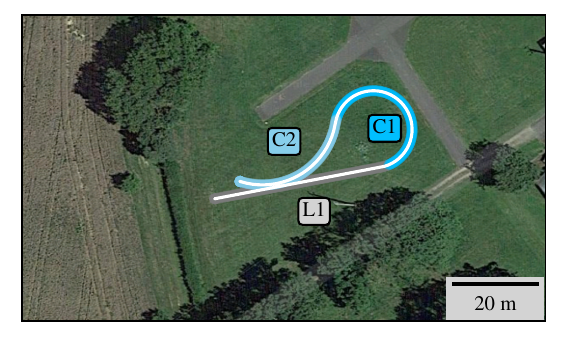}
\caption{Reference path used in the first experiment. It consists of one straight lines $L1$ and two curves ($C1$, $C2$) of different curvatures.}
\label{fig:traj01}
\end{figure}
In this first experiment, several runs were carried out to validate each of the proposed approaches, with the primary objective of enabling the implement point to accurately follow the reference path illustrated in~\autoref{fig:traj01}.
The path was composed of three segments: a straight line ($L1$), a curve with positive curvature ($C1$), and a curve with negative curvature ($C2$).
This configuration was selected to demonstrate convergence and validate the proposed approaches across all curvature categorie, namely, zero, positive, and negative.
In this experiment, two implement positions were considered: one where the implement was located at the rear of the vehicle, and another where it was positioned at the front.
The parameters yielding the best performance for each approach were obtained through manual tuning and are reported in~\autoref{tab:parameters}.
This configuration ensures that each method is compared under its most favorable conditions.
\begin{table}[!t]
    \caption{Table summarizing all the parameters used in the first experiment}
    \label{tab:parameters}
    \centering
    \begin{tabular}{c c c}
        \hline
       &  \textbf{Method}   & \textbf{Parameters}\\
        \hline
        Rear &  Lateral Servoing~\cite{previous_work} & $I_s = \SI{-2.0}{\m}$, $I_y = \SI{-0.5}{\m}$ \\
        \hline
        Rear &  Backstepping~\cite{previous_work}     & $I_s = \SI{-2.0}{\m}$, $I_y = \SI{-0.5}{\m}$ \\
              &                    & $k_y = 0.2$, $k_\theta = 0.6 $ \\
        \hline
        Rear &  Optimal (Our)           & $I_s = \SI{-2.0}{\m}$, $I_y = \SI{-0.5}{\m}$ \\
              &                    & $\lambda = 0.1$, $k_\theta = 0.6 $ \\
              &                    & $s_h = \SI{2.0}{\m}$, $s_t = \SI{0.15}{\m}$ , $h = \SI{2.0}{\s} $\\
        \hline
        Front &  Lateral Servoing~\cite{previous_work} & $I_s = \SI{2.0}{\m}, I_y = \SI{-0.5}{\m}$ \\
        \hline
        Front &  Backstepping~\cite{previous_work}      & $I_s = \SI{2.0}{\m}, I_y = \SI{-0.5}{\m}$ \\
              &                    & $k_y = 0.1, k_\theta = 0.5 $ \\
        \hline
        Front &   Optimal (Our)         & $I_s = \SI{2.0}{\m}, I_y = \SI{-0.5}{\m}$ \\
              &                    & $\lambda = 0.25, k_\theta = 0.3 $ \\
              &                    & $s_h = \SI{1.5}{\m}$, $s_t = \SI{0.15}{\m}$ , $h = \SI{1.5}{\s} $\\
        \hline
    \end{tabular}
\end{table} 
\begin{figure*}[t]
      \centering
       \includegraphics[width=\linewidth]{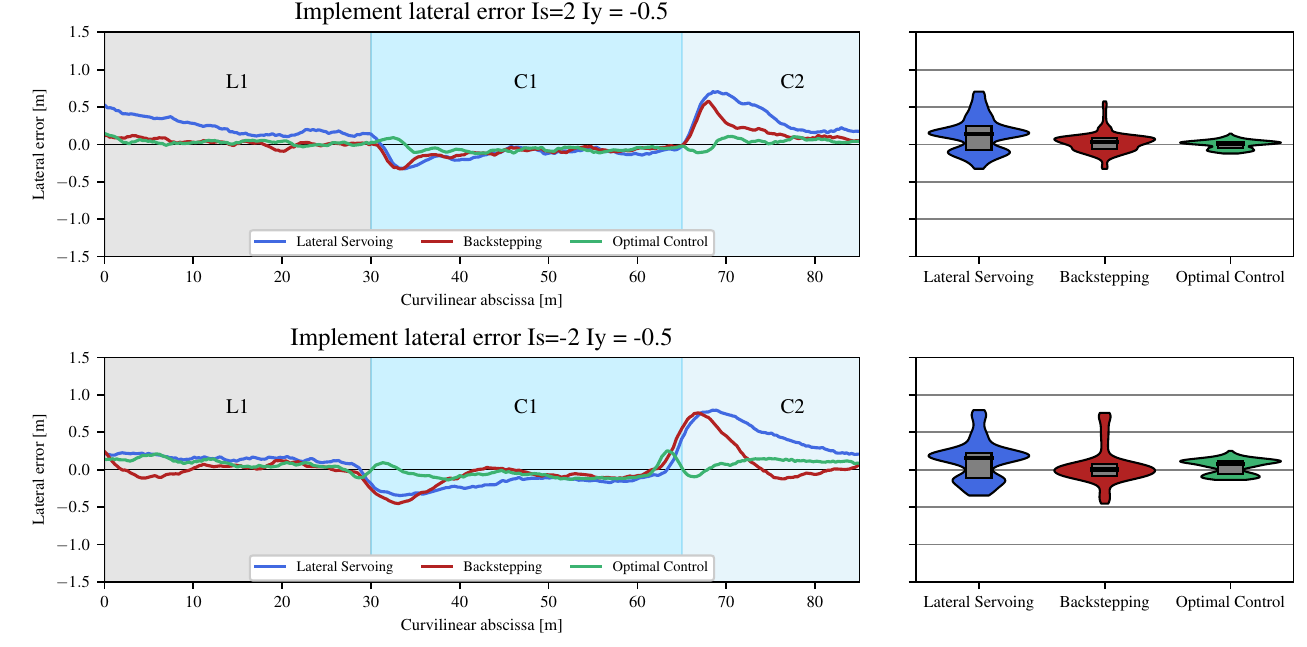}
      \caption{Implement lateral error of a latteral servoing, backstepping and optimal approaches.
      The top graph has the implement at the front of the vehicle, whereas the second graph has the implement at the back. Left: Lateral error as the function of the curvilinear abscissa. The line L1 and curves (C1, C2) correspond to the areas highlighted in \autoref{fig:traj01}. Right: Overall distribution of the errors for each method, with the median and quartiles at 25\% and 75\% represented as box plots.}
      \label{fig:validation}
\end{figure*}

\autoref{fig:validation} presents the lateral deviation error of the implement as a function of the traveled distance. The first plot corresponds to the configuration where the implement is positioned at the front of the vehicle, and the second to the configuration where it is placed at the rear.
As observed with the methods presented in~\cite{previous_work}, the proposed method ensures convergence of the implement point $I$ to the reference path. This is reflected in the lateral error approaching zero.
This convergence is quantitatively confirmed by the median lateral tracking error remaining close to zero, regardless of the implement’s position on the vehicle.
These results indicate that the proposed approach successfully achieves convergence during steady-state phases.

Moreover, significant lateral deviations were observed during curvature transitions when employing the lateral servoing and backstepping methods described in~\cite{previous_work}. These deviations were substantially mitigated by implementing the proposed optimal method. Specifically, the maximum absolue value of the lateral deviation with the optimal approach was approximately $\SI{0.15}{m}$, compared to $\SI{0.6}{m}$ for the alternative methods, representing an improvement of around $\SI{75}{\%}$. Although such overshoots during curvature transitions are inherently unavoidable due to trajectory infeasibility constraints imposed by the implement, the optimal approach significantly enhances lateral accuracy by effectively minimizing the magnitude of deviations.

In conclusion, these experimental results validate the effectiveness and robustness of the proposed optimal method in reducing disturbances, particularly those resulting from curvature changes. This performance improvement stems from the predictive and optimal capabilities of the proposed method, allowing lateral errors to be minimized over the defined prediction horizon.
\subsection{Impact of the prediction horizon}
\begin{figure}[thbp]
      \centering
       \includegraphics[width=\linewidth]{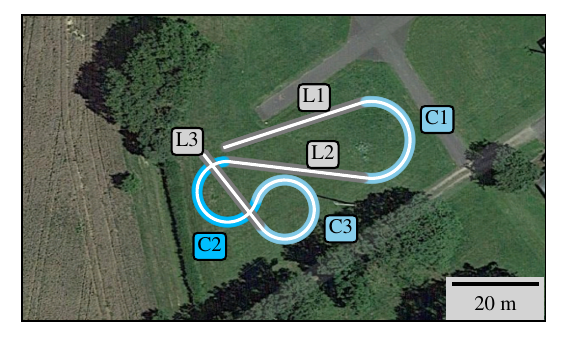}
      \caption{Reference path used in the second experiment. It consists of three straight segments (L1, L2, L3) and three curves (C1, C2, and C3) with varying curvatures. This path includes all types of transition phases: straight-to-curve, curve-to-straight, and curve-to-curve.}
      \label{fig:trajries}
\end{figure}
To evaluate the impact of the prediction horizon on the performance of the proposed method, multiple runs were conducted using various prediction horizon values. The implement point was positioned at the rear of the vehicle, with coordinates $I_s = \SI{-2.0}{\m}$ and $I_y = \SI{-0.5}{\m}$, as this configuration tends to exhibit the highest lateral deviations, as noted in~\cite{previous_work}. The prediction horizon was varied from $\SI{0.5}{\m}$ to $\SI{3.5}{\m}$.
The path used in this experiment is shown in~\autoref{fig:trajries}. It corresponds to a relatively complex path, deliberately designed to introduce multiple disturbances caused by curvature variations. In particular, the path includes transitions between straight and curved segments, as well as between curves of differing curvature.
The presence of several such transitions enables a meaningful assessment of the prediction horizon's effect under dynamic and non-linear conditions. As in the first experiment, manual tuning was performed to determine the parameters that yielded optimal performance for each prediction horizon configuration. These parameters are reported in~\autoref{tab:parameters2}.
\begin{table}[!t]
    \caption{Table summarizing all the parameters used in the second experiment}
    \label{tab:parameters2}
    \centering
    \begin{tabular}{c c}
        \hline
                & \textbf{Parameters}\\
        \hline
        $s_h = 0.5$ &  $\lambda = 0.15$, $k_\theta = 0.35 $, $s_t = \SI{0.10}{\m}$ , $h = \SI{0.5}{\s} $ \\
        \hline
        $s_h = 1.0$ &  $\lambda = 0.15$, $k_\theta = 0.35 $, $s_t = \SI{0.10}{\m}$ , $h = \SI{1.0}{\s} $ \\
        \hline
        $s_h = 1.5$ &  $\lambda = 0.175$, $k_\theta = 0.35 $, $s_t = \SI{0.10}{\m}$ , $h = \SI{1.5}{\s} $ \\
        \hline
        $s_h = 2.0$ &  $\lambda = 0.175$, $k_\theta = 0.4 $, $s_t = \SI{0.10}{\m}$ , $h = \SI{2.0}{\s} $ \\
        \hline
        $s_h = 2.5$ &  $\lambda = 0.2$, $k_\theta = 0.4 $, $s_t = \SI{0.10}{\m}$ , $h = \SI{2.5}{\s} $ \\
        \hline
        $s_h = 3.0$ &  $\lambda = 0.2$, $k_\theta = 0.6 $, $s_t = \SI{0.10}{\m}$ , $h = \SI{3.0}{\s} $ \\
        \hline
        $s_h = 3.5$ &  $\lambda = 0.2$, $k_\theta = 0.6 $, $s_t = \SI{0.10}{\m}$ , $h = \SI{3.5}{\s} $ \\
        \hline
    \end{tabular}
\end{table}

\begin{figure}[t]
      \centering
       \includegraphics[clip,width=\linewidth]{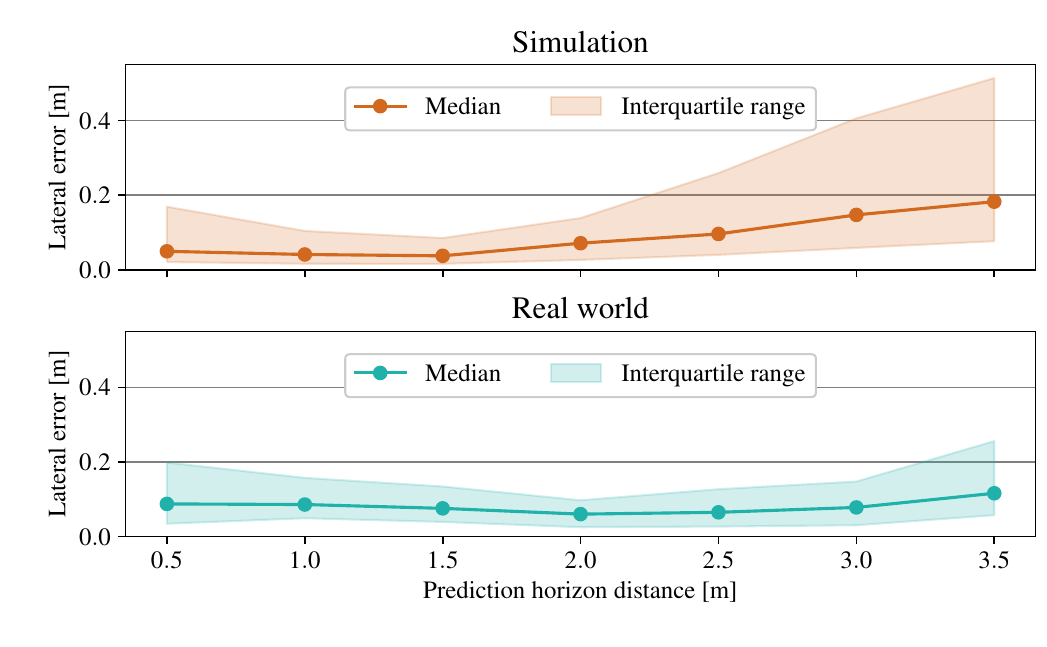}
      \caption{Impact of the prediction horizon on the implement lateral error. The curves represent the medians of the absolute value ot the implement lateral deviation, the colored area the interquartile range between 25\% and 75\%.}
      \label{fig:impact_of_sh}
\end{figure}

Before conducting field experiments, simulations were performed to anticipate the expected system behavior. The first plot in~\autoref{fig:impact_of_sh} presents the results of these simulations, showing the median tracking error along with the corresponding interquartile range for each selected prediction horizon.

From these simulation results, the existence of a minimum can be observed, suggesting that there exists an optimal prediction horizon that improves path tracking. Increasing the prediction horizon enhances the controller’s ability to anticipate future disturbances. However, if the horizon becomes too large, it may lead to a degradation in tracking performance. In such cases, the control law tends to focus on minimizing errors caused by distant disturbances, which may come at the cost of short-term accuracy, ultimately leading to a degradation in overall tracking performance.

Following this simulation-based analysis, field experiments were conducted, yielding the second plot in~\autoref{fig:impact_of_sh}.
A strong correlation was observed between the simulation and experimental results, confirming the reproducibility of the findings. As in simulation, the experimental curves exhibit the presence of an optimal prediction horizon that minimizes the implement’s lateral deviation and, consequently, the overall tracking error thus validating the earlier analysis.

These results highlight the critical influence of the prediction horizon on the behavior of the control law. While an appropriately selected horizon allows for effective anticipation of future errors and improved tracking performance, excessive anticipation over long distances may degrade local accuracy by overemphasizing distant disturbances.
In this study, the optimal prediction horizon was identified as $\SI{2.0}{\m}$, corresponding to the longitudinal offset of the implement.

\section{Conclusion}
In this paper, an optimal control strategy was proposed for accurately tracking an offset point rigidly attached to a mobile robot. Precisely tracking such a point, typically located on an implement, is a crucial requirement in agricultural robotics, as the implement directly performs the primary agronomic operations.

It was demonstrated that previously developed lateral deviation servoing and backstepping methods presented in~\cite{previous_work} may exhibit undesirable overshoot during transient phases. This behavior can be particularly detrimental in agricultural applications, where precision is critical to avoid crop damage.
To overcome this limitation, an optimal control approach was introduced. The proposed method ensures accurate convergence of the implement's control point onto the reference path while significantly mitigating overshoot during transient responses.

Future work will focus on incorporating dynamic effects such as wheel slip and extending the optimal control framework to handle multiple points on the robot, including wheel-soil contact points and various implement positions. Such advancements are essential not only to preserve precise implement positioning but also to ensure that the robot wheels avoid encroaching upon crop rows, thereby minimizing potential crop damage.






\section*{ACKNOWLEDGMENT}
This work has been funded by the french National Research Agency (ANR), under the grant ANR-19-LCV2-0011, attributed to the joint laboratory Tiara (\url{www6.inrae.fr/tiara}). It has also received the support of the French government research program "Investissements d'Avenir" through the IDEX-ISITE initiative 16-IDEX-0001 (CAP 20-25), the IMobS3 Laboratory of Excellence (ANR-10-LABX-16-01).  This work has been partially supported by ROBOTEX 2.0 (Grants ROBOTEX ANR-10-EQPX-44-01 and TIRREX ANR-21-ESRE-0015)


\printbibliography

\end{document}